%% file: main.tex
\pdfoutput=1

\documentclass[11pt]{article}

\usepackage{naacl2021}

\usepackage{times}
\usepackage{latexsym}
\usepackage{cancel}
\usepackage{graphicx}
\usepackage{amsmath}
\usepackage{latexsym}
\usepackage{amssymb}
\usepackage{amsthm}
\usepackage{color}
\usepackage{xspace}
\usepackage{subfigure}
\usepackage{array, booktabs, makecell}

\usepackage{booktabs}
\usepackage{array,tabularx}
\usepackage{ulem}
\usepackage{amssymb}
\usepackage{pifont}
\usepackage{multirow}
\newenvironment{conditions*}
  {\par\vspace{\abovedisplayskip}\noindent
   \tabularx{\columnwidth}{>{$}l<{$} @{}>{${}}c<{{}$}@{} >{\raggedright\arraybackslash}X}}
  {\endtabularx\par\vspace{\belowdisplayskip}}

\usepackage{microtype}

\input{commands.tex}

\title{Neural Language Modeling for Contextualized Temporal Graph Generation}

\author{Aman Madaan, Yiming Yang \\
  Language Technologies Institute, Carnegie Mellon University \\
  Pittsburgh, PA, USA \\
  \texttt{amadaan@cs.cmu.edu} \\}

\date{}

\begin{document}
\maketitle

\input{introduction.tex}
\input{relatedwork.tex}
\input{datacreation.tex}
\input{model.tex}
\input{experiments.tex}
\input{conclusion-future.tex}

\bibliography{main}
\bibliographystyle{acl_natbib}
\input{appendix.tex}
\end{document}

%% file: commands.tex
\newcommand{\caevo}{\textsc{caevo}\xspace}
\newcommand{\nyt}{\textsc{nyt}\xspace}
\newcommand{\before}{\textsc{before}\xspace}

\newcommand{\ours}{\textsc{gpt-2}\xspace}
\newcommand{\dotlang}{\textsc{dot}\xspace}
\newcommand{\bleu}{\textsc{bleu}\xspace}
\newcommand{\ged}{\textsc{ged}\xspace}
\newcommand{\sts}{Seq2Seq\xspace}
\newcommand{\cct}{Cogcomptime\xspace}
\newcommand{\gptz}{\textsc{gpt-2}\xspace}
\newcommand{\after}{\textsc{after}\xspace}
\newcommand{\incld}{\textsc{includes}\xspace}
\newcommand{\isincluded}{\textsc{is included}\xspace}
\newcommand{\simul}{\textsc{simultaneous}\xspace}

\newcommand{\tggen}{TG-Gen\xspace}
\newcommand{\tbden}{TB-Dense\xspace}
\definecolor{commred}{RGB}{244,144,151}
\definecolor{commgreen}{RGB}{164,194,151}
\definecolor{commyellow}{RGB}{245,233,96}
\definecolor{commblue}{RGB}{223,178,244}
\definecolor{gray}{RGB}{220,220,220}

\def\V#1{\mathbf{#1}}

\def\S#1{\boldsymbol{#1}}

%% file: introduction.tex
\begin{abstract}
This paper presents the first study on using large-scale pre-trained language models for automated generation of an event-level temporal graph for a document. Despite the huge success of neural pre-training methods in NLP tasks, its potential for temporal reasoning over event graphs has not been sufficiently explored. Part of the reason is the difficulty in obtaining large training corpora with human-annotated events and temporal links. We address this challenge by using existing IE/NLP tools to automatically generate a large quantity (89,000) of system-produced document-graph pairs, and propose a novel formulation of the contextualized graph generation problem as a sequence-to-sequence mapping task. These strategies enable us to leverage and fine-tune pre-trained language models on the system-induced training data for the graph generation task. Our experiments show that our approach is highly effective in generating structurally and semantically valid graphs. Further, evaluation on a challenging hand-labeled, out-of-domain corpus shows that our method outperforms the closest existing method by a large margin on several metrics.
We also show a downstream application of our approach by adapting it to answer open-ended temporal questions in a reading comprehension setting.\footnote{Code and pre-trained models available at~\url{https://github.com/madaan/temporal-graph-gen}}
\end{abstract}

\section{Introduction}
\begin{figure*}[ht]
\centering
    {\includegraphics[width=0.95\textwidth,height=0.35\textheight]{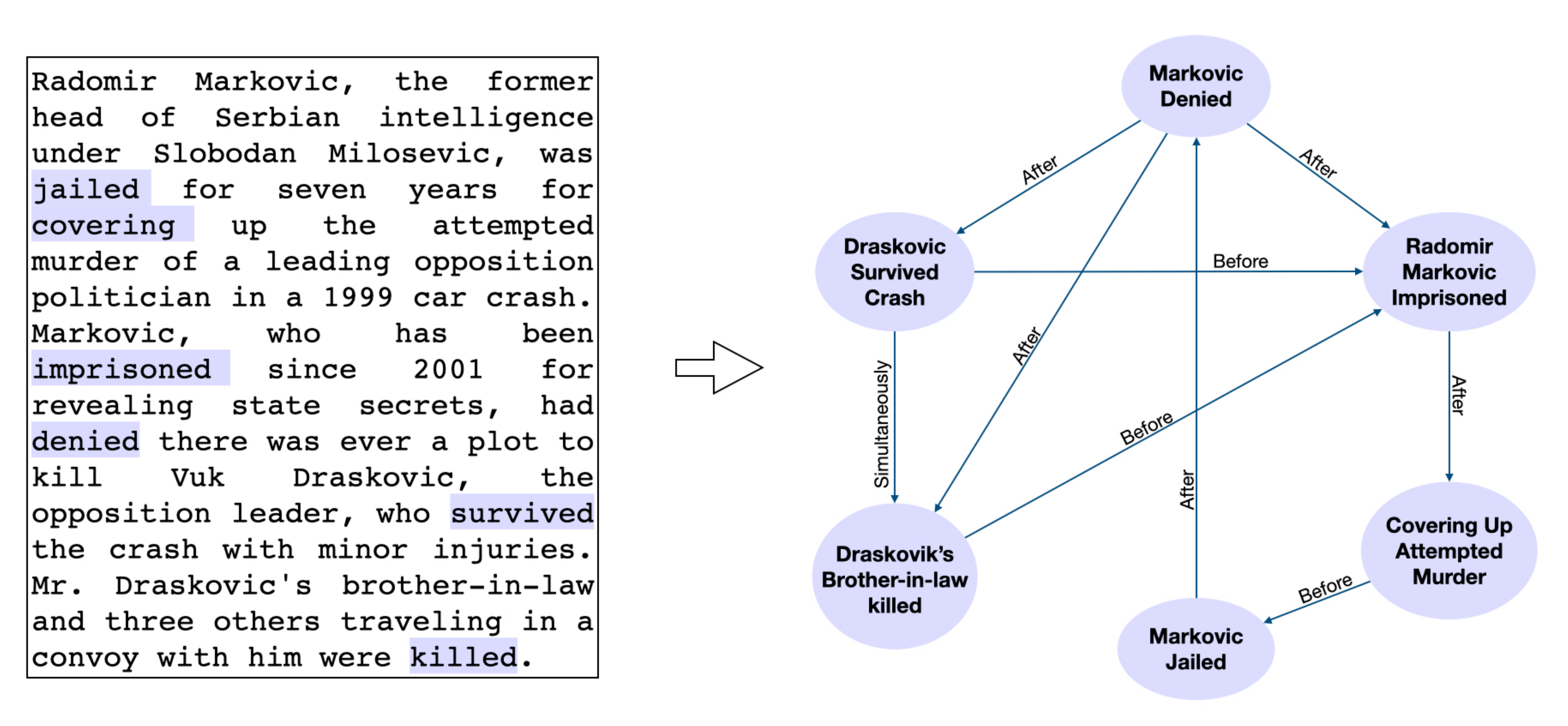}} 
    \caption{Task overview: given a document~(left), automatically extract a temporal graph~(right).} 
    \label{fig:task}
\end{figure*}
 
Temporal reasoning is crucial for analyzing the interactions among complex events and producing  
coherent interpretations of text data ~\cite{duran2007using}.
There is a rich body of research on the use of temporal information in a variety of important application domains, including topic detection and tracking~\cite{makkonen2003topic}, information extraction~\cite{ling2010temporal}, parsing of clinical records ~\cite{lin2016multilayered}, discourse analysis~\cite{evers2017temporality}, and question answering~\cite{ning2020torque}.

Graphs are a natural choice for representing the temporal ordering among events, where the nodes are the individual events, and the edges capture temporal relationships such as ``before'', ``after'' or ``simultaneous''.
Representative work on automated extraction of such graphs from textual documents includes the early work by~\citet{chambers2009unsupervised}, where the focus is on the construction of event chains from a collection of documents, and the more recent \caevo~\cite{chambers2014dense} and
\cct~\cite{ning2018cogcomptime}, which extract a graph for each input document instead.  
These methods focus on rule-based and statistical sub-modules to extract verb-centered events and the temporal relations among them.
As an emerging area of \textsc{nlp}, large scale pre-trained language models have made strides in addressing challenging tasks like commonsense knowledge graph completion~\cite{Bosselut2019COMETCT} and task-oriented dialog generation~\cite{budzianowski2019hello}.
These systems typically fine-tune large language models on a corpus of a task-specific dataset.
However, these techniques have not been investigated for temporal graph extraction.

This paper focuses on the problem of generation of an event-level temporal graph for each document, and we refer to this task as \textit{contextualized} graph generation.  
We address this open challenge by proposing a novel reformulation of the task as a sequence-to-sequence mapping problem~\cite{sutskever2014sequence}, which enables us to leverage large pre-trained models for our task.
Further, different from existing methods, our proposed approach is completely end-to-end and eliminates the need for a pipeline of sub-systems commonly used by traditional methods.

We also address a related open challenge, which is a prerequisite to our main goal: the difficulty of obtaining a large quantity of training graphs with human-annotated events and temporal relations.  
To this end, we automatically produce a large collection of document-graph pairs by using \caevo, followed by a few rule-based post-processing steps for pruning and noise reduction. 
We then encode the graph in each training pair as a string in the graph representation format \dotlang, transforming the text-to-graph mapping into sequence-to-sequence mapping. 
We fine-tune \gptz on this dataset of document-graph pairs, which yields large performance gains over strong baselines on system generated test set and outperforms \caevo on TimeBank-Dense~\cite{cassidy2014annotation} on multiple metrics.
Figure 1 shows an example of the input document and the generated graph by our system.
In summary, our main contributions are:
\begin{enumerate}
\itemsep0em 
\item We present the first investigation on using large pre-trained language models for contextualized temporal event graph generation by proposing a new formulation of the problem as a sequence-to-sequence mapping task.
\item 
We address the difficulty of obtaining a large collection of human-annotated graphs, which is crucial for effective fine-tuning of pre-trained models, by automatically producing a collection of 89,000 document-graph pairs.
\item Our experimental results on both the system-generated test set (which allows us to compare the relative performance of different models) and a hand-labeled, out-of-domain dataset~(TimeBank-Dense),
show the advantage of our proposed approach over strong baselines.
Further, we show that our approach can help in generating plausible answers for open ended-temporal questions in a reading comprehension dataset, Torque~\cite{ning2020torque}.
\end{enumerate}

%% file: relatedwork.tex
\section{Related Work}

\paragraph{Temporal Graph Extraction }
Tempeval-3~\cite{uzzaman2013semeval} introduced the task of temporal graph extraction as ``the ultimate task for evaluating an end-to-end system that goes from raw text to TimeML annotation''.
Notable systems developed in response include \caevo~\cite{chambers2014dense}, followed by the more recent \cct~\cite{ning2018cogcomptime}. 
Both \caevo and \cct use several statistical and rule-based methods like event extractors, dependency parsers, semantic role labelers, and time expression identifiers for the task.
Our work differs from these systems in both the methodology and desired result in the following ways: i) Instead of using specialized sub-systems, we transform the task into a sequence-to-sequence mapping problem and use a single language model to generate such temporal graphs in an end-to-end fashion from text, subsuming all the intermediate-steps.
ii) We develop our system using a corpus of 89,000 documents, which is $\sim$ 300x larger compared to datasets used by \caevo~(36 documents) and \cct on~(276 documents);
iii) We remove the noisy events included by \caevo, but do not limit the extracted events to any specific semantic axis as done by \cct; and finally,
iv) Our method generates graphs where the nodes are not simple verbs but augmented event phrases, containing the subject and the object of each verb.
We use \caevo over \cct to generate a large-scale corpus for our task and to evaluate our system for the following reasons: i) We found \caevo to be much more scaleable, a critical feature for our task of annotating close to 100k documents, ii) \caevo over-generates (and not excludes) verbs from its output, giving us the flexibility to filter out noisy events without inadvertently missing out on any critical events.
However, our method makes no assumption specific to \caevo and is adaptable to any other similar system (including \cct).

\paragraph{Temporal relation extraction}
We note that the problem of temporal graph extraction is different from the more popular task of Temporal relation extraction (Temprel), which deals with classifying the temporal link between two already extracted events.
State of the art Temprel systems use neural methods~\cite{ballesteros2020severing,ning2019improved,goyal-durrett-2019-embedding,han2019deep,cheng2017classifying}, but typically use a handful of documents for their development and evaluation.
\citet{vashishtha-etal-2019-fine} are a notable exception by using Amazon Mechanical Turks to obtain manual annotations over a larger dataset of 16,000 sentences.
We believe that the techniques presented in our work can be applied to scale the corpus used for training Temprel systems.

\paragraph{Language Models for Graph Generation }
Recently, \citet{Bosselut2019COMETCT} proposed \textsc{comet}, a system that fine-tunes \textsc{gpt}~\citep{radford2018improving} on commonsense knowledge graphs like \textsc{atomic}~\cite{sap2019atomic} and conceptnet~\citep{speer2017conceptnet} for commonsense kb completion.
Similar to~\textsc{comet}, we adopt large-scale language models for such a conditional generation of text.
However, our task differs from \textsc{comet} in the complexity of both the conditioning text and generated text: we seek to generate temporal graphs grounded in a document, whereas \textsc{comet} generates a short event/concept phrase conditioned on a relation and an input event/concept phrase.
\citet{madaan2020eigen} and \citet{rajagopal2021curie} aim to generate event influence graphs grounded in a situation.
Similar to this work, these methods rely on pre-trained language models to generate informative structures grounded in text.
Different from us, these methods break the generation process into a sequence of natural language queries.
Each query results in an event node, which are finally assembled into a tree.
In contrast, we propose a method to directly generate graphs with arbitrary topology from text.
Additionally, the events generated by these methods are not present in text making event event prediction, rather than event extraction as their primary focus.
\citet{you2018graphrnn} formulate graphs as a sequence for learning generative models of synthetic and real-world graphs.
Similar to their work, we formulate graph generation as an auto-regressive task.
However, our goal is the conditional generation of temporal graphs, and not learning unconditional generative distributions.
Finally, inspired by recent trends~\cite{raffel2019exploring}, we do not make any graph specific modifications to the model or the decoding process and formulate the problem as a straightforward sequence-to-sequence mapping task.
While our approach does not rely on any particular language model, it would be interesting to see the gains achieved by the much larger \textsc{gpt-3}~\cite{brown2020language} on the dataset produced by our method.\footnote{Not available for research as of April 2021.}


%% file: datacreation.tex
\section{Deriving Large-scale Dataset for the Temporal Graph Generation}
\label{sec:dataset}

\noindent{\bf Definitions and Notations: }
Let $\V{G}(\V{V}, \V{E})$ be a temporal graph associated with a document $\V{D}$, such that vertices $\V{V}$ are the events in document $\V{D}$, and the edges $\V{E}$ are temporal relations (links) between the events.
Every temporal link in $\V{E}$ takes the form $r(e_q, e_t)$ where the query event $e_q$ and the target event $e_t$ are in $\V{V}$, and $r$ is a temporal relation (e.g., before or after). 
In this work, we undertake two related tasks of increasing complexity: i) Node generation, and ii) Temporal graph generation:

\vspace{0.3em}
\noindent{\bf Task 1: Node Generation: }
\label{sec:task1}
\textit{Let $r(e_q, e_t)$ be an edge in $\V{E}$.
Let $C_r$ be the set of sentences in the document $\V{D}$ that contains the events $e_q$ or $e_t$ or are adjacent to them.
Given a query consisting of $C_r$, $r$, and $e_q$, generate $e_t$.}


\vspace{0.3em}
\noindent{\bf Task 2: Temporal Graph Generation: }
\label{sec:task2}
\textit{Given a document $\V{D}$, generate the corresponding temporal graph $\V{G}(\V{E}, \V{V})$.}

Figure~\ref{fig:task} illustrates the two tasks.
Task 1 is similar to knowledge base completion, except that the output events $e_q$ are generated, and not drawn from a fixed set of events.
Task 2 is significantly more challenging, requiring the generation of both the structure and semantics of $\V{G}$.

The training data for both the tasks consists of tuples $\{(x_i, y_i)\}_{i=1}^{N}$.
For Task 1, $x_i$ is the concatenation of the query tokens $(C_r, e_q, r)$, and $y_i$ consists of tokens of event $e_t$.
For Task 2, $x_i$ is the $i^{th}$ document $\V{D_i}$, and $y_i$ is the corresponding temporal graph $\V{G_i}$.

We use the New York Times (\nyt) Annotated Corpus \footnote{https://catalog.ldc.upenn.edu/LDC2008T19} to derive our dataset of document-graph pairs. 
The corpus has 1.8 million articles written and published by \nyt between 1987 and 2007.
Each article is annotated with a hand-assigned list of descriptive terms capturing its subject(s).
We filter articles with one of the following descriptors: \{``bomb'', ``terrorism'', ``murder'', ``riots'', ``hijacking'', ``assassination'', ``kidnapping'', ``arson'', ``vandalism'', ``hate crime'', ``serial murder'', ``manslaughter'', ``extortion''\}, yielding 89,597 articles, with a total of 2.6 million sentences and 66 million tokens. 
For each document $\V{D}$, we use \caevo~\cite{chambers2014dense} to extract the dense temporal graph consisting of i) the set of verbs, and ii) the set of temporal relations between the extracted verbs. \caevo extracts six temporal relations: before, after, includes, is included, simultaneous, and vague.

We process each dense graph extracted by \caevo with a series of pruning and augmentation operations: \textbf{i)} We observed that some of the most frequent verbs extracted by \caevo were the so-called reporting verbs~\cite{liu2018automatic}, like \textit{said}, \textit{say}, and \textit{
told}, which do not contribute to the underlying events.
For example, \textit{said} formed nearly 10\% of all the verbs extracted by \caevo as an event.
To remove such noisy events, we remove the five verbs with the lowest inverse document frequencies, as well as an additional set of light and reporting verbs~\cite{liu2018automatic,recasens2010typology}\footnote{The final list of verbs is: i) \textit{low idf:} ``said'', ``say'', ``had'', ``made'', ``told'',
ii) \textit{light:} ``appear'', ``be'', ``become'', ``do'', ``have'', ``seem'', ``get'', ``give'', ``go'', ``have'', ``keep'', ``make'', ``put'', ``set'', ``take'',
iii) \textit{reporting:} ``argue'', ``claim'', ``say'', ``suggest'', ``tell''.}
\textbf{ii)} To make event annotations richer, we follow~\cite{chambers2008unsupervised}, and prefix and suffix every verb with its noun-phrase and object, respectively.
This augmentation helps in adding a context to each verb, thus making events less ambiguous.
For instance, given a sentence: \textit{A called B, after which B called C},
\caevo extracts $\after(\textit{called}, \textit{called})$.
With the proposed augmentation, the relation becomes $\after(\textit{A called B}, \textit{B called C})$, clearly differentiating the two different \textit{called} events.
Our notion of events refers to such augmented verbs.
Crucially, different from prior work, our system is trained to extract these augmented event phrases.
We also drop all the verbs that do not have either a subject or an object.
\textbf{iii)} We remove the relations extracted by the statistical sieves if they have a confidence score of less than 0.50 and retain the rule-based relations as those were shown to be extracted with a high precision by~\citet{chambers2014dense}.
Finally, we only retain event-event relations (dropping links between verbs and time expressions) and drop the vague relations as they typically do not play any role in improving the understanding of the temporal sequences in a document.
As Table~\ref{tab:pruning} shows, pruning noisy verbs and relations yields sparser and more informative graphs.
\begin{table}[ht]
\setlength{\tabcolsep}{0.25em}
\centering
\begin{tabular}{@{}lrrr@{}}
\toprule
                & Initial  & Pruned & \% Reduction\\ \midrule
\#Relations & 27,692,365 & 4,469,298 & 83.86 \\
\#Events & 6,733,396 & 2,615,296 & 61.15 \\\bottomrule
\end{tabular}
\caption{Effect of pruning operations on the number of relations and events.
}
\label{tab:pruning}
\end{table}

\paragraph{Creating Sub-graphs using Event Communities}
We discovered that the (pruned) graph generated for a given document typically has several sub-graphs that are either completely disconnected or have high intra-link density.
Further, we found that each of these sub-graphs are grounded in different parts of the document.
We exploit this phenomenon to map each sub-graph to its correct context, thus reducing the noise in the data.

Relying merely on connectivity for creating sub-graphs is still prone to noise, as largely unrelated sub-graphs are often connected via a single event.
Instead, we propose a novel approach based on the detection of event communities to divide a graph into sub-graphs, such that the events in a sub-graph are more densely connected to each other.
We learn these event communities using the concept of modularity, first introduced by~\cite{newman2004finding}.
We defer the derivation of modularity optimization to the Appendix.

\label{sec:modularity}
\vspace{0.5em}
\noindent{\bf Datasets for Task 1 and Task 2}
After running the pruning and clustering operations outlined above on 89k documents, we obtain a corpus of over 890,677 text-graph pairs, with an average of 120.31 tokens per document, and 3.33 events and 4.91 edges per graph.
These text-graph pairs constitute the training data for Task 2.
We derive the data for Task 1 from the original (undivided) 89k graphs~(each document-graph pair contributes multiple examples for Task 1).
In Task 1 data, nearly 80\% of the queries $(C_r, e_q, r)$ had a unique answer $e_t$, and nearly 16\% of the queries had two different true $e_t$.
We retain examples with multiple true $e_t$ in the training data because they help the model learn diverse temporal patterns that connect two events.
For fairness, we retain such cases in the test set.
Table~\ref{tab:data-stats} lists the statistics of the dataset.
The splits were created using non-overlapping documents.


\begin{table}[ht]
\centering
\begin{tabular}{lrrr}
\toprule
Task   & train & valid & test  \\ \midrule
Task 1 & 4.26 & 0.54 & 0.54  \\  \midrule
Task 2 & 0.71 & 0.09 & 0.09 \\ \bottomrule
\end{tabular}
\caption{Dataset statistics (counts in million).}
\label{tab:data-stats}
\end{table}

\subsection{Graph Representation}
\begin{figure}[ht]
\centering
    {\includegraphics[width=0.46\textwidth]{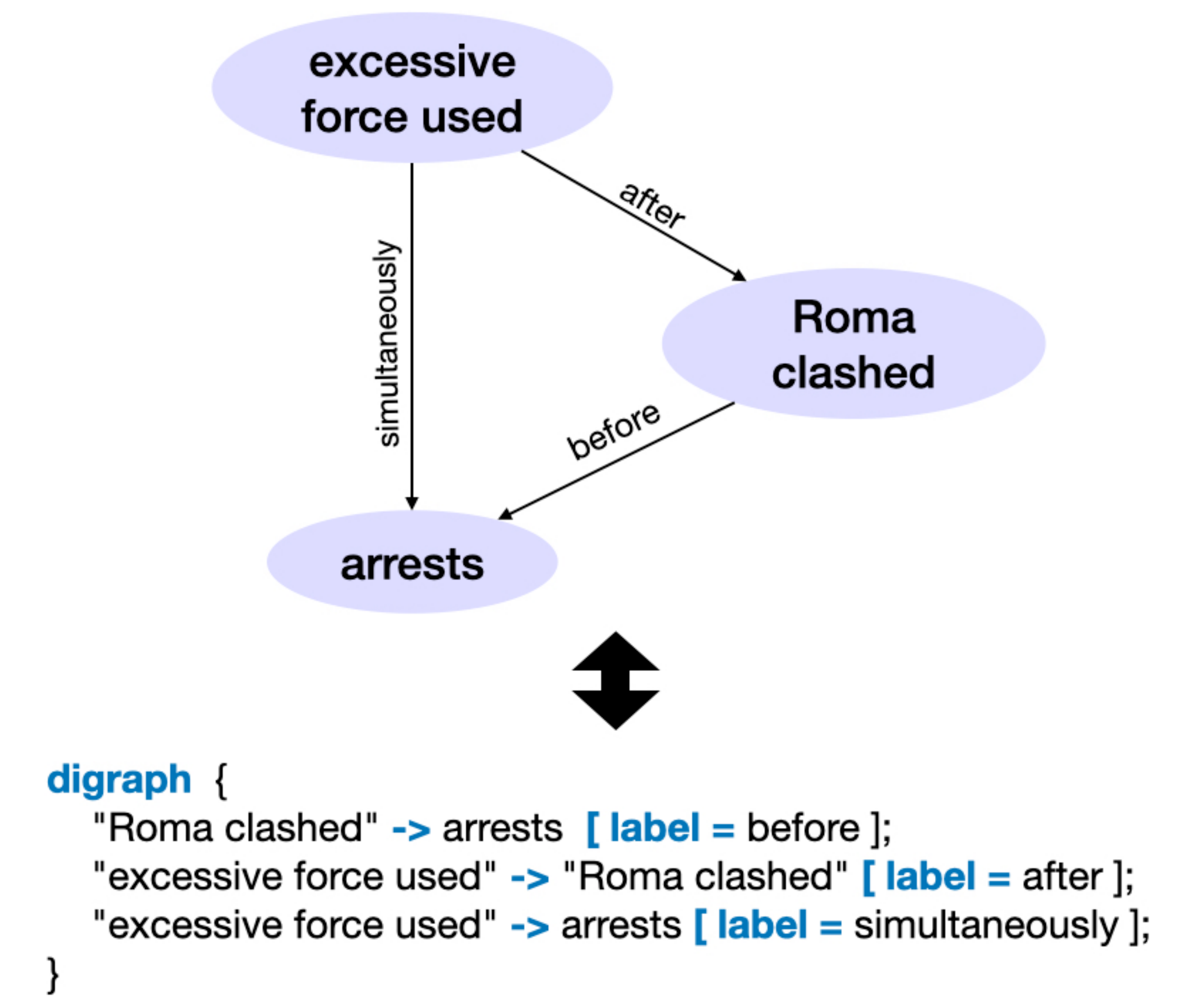}}
    \caption{Temporal graph and the corresponding \dotlang representation for the sentence: \textit{Roma clashed fiercely with the police, leading to arrests in which Roma activists said excessive force was used.}} 
    \label{fig:dot-example}
\end{figure}

We use language models to generate each graph as a sequence of tokens conditioned on the document, thus requiring that the graphs are represented as strings.
We use \dotlang language~\cite{gansner2006drawing} to format each graph as a string.
While our method does not rely on any specific graph representation format, we use \dotlang as it supports a wide variety of graphs and allows augmenting graphs with node, edge, and graph level information.
Further, graphs represented in \dotlang are readily consumed by popular graph libraries like NetworkX~\cite{SciPyProceedings_11}, making it possible to use the graphs for several downstream applications.
Figure~\ref{fig:dot-example} shows an example graph and the corresponding \textsc{dot} code.
The edges are listed in the order in which their constituent nodes appear in the document.
This design choice was inspired by our finding that a vast majority of temporal links exist between events that are either in the same or in the adjoining sentence (this phenomenon was also observed by~\citet{ning2019structured}).
Thus, listing the edges in the order in which they appear in the document adds a simple inductive bias of locality for the auto-regressive attention mechanism, whereby the attention weights \textit{slide} from left to right as the graph generation proceeds.
Additionally, a fixed order makes the problem well defined, as the mapping between a document and a graph becomes deterministic.

%% file: model.tex
\section{Model}
\label{sec:model}
The training data $\V{X}$ for both Tasks 1 and 2 comprises of tuples $\{(\S{x_i}, \S{y_i})\}_{i=1}^{N}$. 
For task 1 (node generation), $\S{x_i}$ the concatenation of context, the source, node, and the relation.
The target $\S{y_i}$ consists of the tokens of the target event.
For task 2 (graph generation), $\S{x_i}$ is a document and $\S{y_i}$ is the corresponding temporal graph represented in \dotlang.
We train a (separate) conditional language model to solve both the tasks.
Specifically, given a training corpus of the form $\{(\S{x_i}, \S{y_i})\}$, we aim to estimate the distribution $p_{\theta}(\S{y_i} \mid \S{x_i})$. 
Given a training example $(\S{x_i}, \S{y_i})$ we set $\S{u_i} = \S{x_i} \| \S{y_i}$\footnote{$\|$ denotes concatenation}.
$p_\theta(\S{u_i})$ can then be factorized as a sequence of auto-regressive conditional probabilities using the chain rule:  $p_\theta(\S{u_i}) = \prod_{k=1}^{n} p (u_{i,k} | \S{u_{i, <k}}) $, where $u_{i,k}$ denotes the $k^{th}$ token of the $i^{th}$ sequence, and $\S{u_{i, <k}}$ denotes the sequence of tokens $\{u_1, u_2, ..., u_{k - 1}\}$.
Language models are typically trained by minimizing a cross-entropy loss $-\text{log}p_\theta(\S{u_i})$ over each sequence $\S{u_i}$ in $\V{X}$.
However, the cross-entropy loss captures the joint distribution $p_\theta(\S{x_i}, \S{y_i})$, and is not aligned with our goal of learning conditional distribution $p_\theta(\S{y_i} \mid \S{x_i})$.
To circumvent this, we train our model by masking the loss terms corresponding to the input $\S{x_i}$, similar to \citet{Bosselut2019COMETCT}.
Let $\S{m_i}$ be a mask vector for each sequence $\S{u_i}$, set to $0$ for positions corresponding to $\S{x_i}$, and $1$ otherwise i.e. $m_{i, j} = 1$ $\text{if } j > |\S{x_i}|$, else 0.
We combine the mask vector with our factorization of $p_\theta\left(\S{u_i}\right)$ to formulate a \textit{masked} language modeling loss $\mathcal{L}$, which is minimized over the training corpus $\V{X}$ to estimate the optimal $\theta$:
$$
        \mathcal{L}(\V{X}) = - \sum_{i=1}^{|\V{X}|} \sum_{j=1}^{|x_i| + |y_i|} m_{i, j}  * \log\left(p_\theta \left(u_{i,j} | \S{u_{i, <j}}\right)\right) 
$$
Note that the formulation of masked loss is opaque to the underlying architecture, and can be implemented with a simple change to the loss function.
In practice, we use \gptz~\cite{radford2019language} based on transformer architecture~\cite{vaswani2017attention} for our implementation.
Having trained a $p_{\theta}$ for each task, we generate a node~($\S{y}$) given a query~($\S{x}$) (for Task 1), or a graph~($\S{y}$) given a document~($\S{x}$) (for Task 2) by drawing samples from the appropriate $p_{\theta}(\S{y} \mid \S{x})$ using nucleus sampling~\cite{holtzman2019curious}.
We provide more details of our training procedure and the architecture in the Appendix~(\ref{sec:gpt2mlm}).

%% file: experiments.tex
\section{Experiments and Results}

\begin{table}
\centering
\setlength{\tabcolsep}{0.29em}
\begin{tabular}{@{}llrrrr@{}}
\toprule
Method & Dataset & \bleu  & \textsc{mtr} & \textsc{rg} & \textsc{acc}  \\ \midrule
\sts  & {\small \tggen(-C)} & 20.20 & 14.62 & 31.95 & 19.68 \\
\sts  & {\small \tggen} & 21.23 & 16.48 & 35.54 & 20.99 \\
\ours & {\small \tggen(-C)}  & 36.60 & 25.11  & 43.07 & 35.07 \\
\ours  & {\small \tggen} & \textbf{62.53} & \textbf{43.78}  & \textbf{69.10} & \textbf{61.35}\\  \midrule
\sts  & {\small \tbden(-C)} & 11.55 & 9.23   & 21.87 & 10.06 \\
\sts  & {\small \tbden} & 16.68 & 12.69   & 27.75 & 13.97  \\
\ours  & {\small \tbden(-C)} & 22.35 & 15.04  & 27.73 & 20.81 \\
\ours  & {\small \tbden} & \textbf{52.21} & \textbf{35.69}   & \textbf{57.98} & \textbf{47.91} \\  \bottomrule
\end{tabular}
\caption{Node Generation (task 1) results.
}
\label{tab:task1-automated-metrics}

\end{table}

\subsection{Evaluation Datasets}
We evaluate our method on two different datasets: i) \textbf{\tggen}: Test split of synthetically created dataset~(Section \ref{sec:dataset}), and ii) \textbf{\tbden:} A mixed-domain corpus, with human-annotated temporal annotations.
We create \tbden from the test splits of  TimeBank-Dense~\cite{cassidy2014annotation} by applying the same pre-processing operations as we did for \tggen.
\tbden forms a very challenging dataset for our task because of domain mismatch; our system was trained on a corpus of terrorism-related events, whereas \tbden includes documents from a wide array of domains, forming a zero-shot evaluation scenario for our method.

\subsection{Implementation Setup}
\noindent \textbf{\ours: }
We use \textsc{gpt-2} medium (355M parameters) for our experiments with 24-layers, a hidden size of 1024, and 16 self-attention heads.
We build on the implementation by~\citet{wolf2019huggingface}, using the default hyperparameters and a block size input sequence length after tokenization) of 512.
For optimization, we use AdamW~\cite{loshchilov2017decoupled} with a learning rate of 5e-5, a batch size of 1, and no learning rate scheduling.
We also experimented with a block size of 300 and a batch size of 2.
We found the results (presented in the appendix) to be worse, underscoring the importance of using a larger block size for generating larger outputs.
We generate samples using nucleus sampling using $p=0.9$, and set maximum output length to be 500 for graph generation and 60 foe node generation.
All of our experiments were done on a single Nvidia GeForce \textsc{rtx} 2080 Ti.
The models were initialized with the pre-trained weights provided by~\citet{radford2019language}, and fine-tuned for five epochs, with a run-time of 48 hours/epoch for Task 1 and 52 hours/epoch for Task 2.
We use the last checkpoint (i.e., at the end of fine-tuning) for all experiments. 
Despite the higher perplexity on the development set, we found the overall performance of the last checkpoint to be better.

We also experimented with \gptz without fine-tuning (i.e., by directly using pre-trained weights).
The non-finetuned \gptz fared poorly for both the tasks across all the metrics, getting a \bleu score of near 0 for Task 1.
This dismal performance underscores the importance of fine-tuning on the end task for large-scale pre-trained language models.

Finally, we note that our method does not make any model-specific assumption, and can be used with any auto-regressive language model (i.e., a language model that generates a sequence of tokens from left-to-right).
We use \gptz as a representative for large pre-trained language models.

\vspace{0.5em}
\noindent \textbf{\sts: } We train a bi-directional \textsc{lstm}~\cite{hochreiter1997long} based sequence-to-sequence model~\cite{bahdanau2015neural} with global attention~\cite{luong2015effective} and a hidden size of 500 as a baseline to contrast with \gptz.
The token embeddings initialized using 300-dimensional pre-trained Glove~\cite{pennington2014glove}.


\subsection{Task 1: Node Generation}
\begin{table}[ht]
\setlength{\tabcolsep}{0.15em}
\small{
\begin{tabular}{p{7.5cm}}
\toprule
\textbf{Paragraph:} \textit{Mr. Grier, a former defensive lineman for the New York Giants who was \textcolor{commred}{\textbf{ordained}} as a minister in 1986, \textcolor{commgreen}{\textbf{testified}} on Dec. 9 that he had \textbf{visited} Mr. Simpson a month earlier} \\ \midrule
\end{tabular}
}

\caption{An example of \gptz \textit{fixing} the label given by \caevo. Given a query \textit{event \textbf{after} ``Mr. Grier visited''}, \caevo incorrectly extracts \textit{Mr. Grier ordained}, whereas \gptz generates the correct event: \textit{Mr. Grier testified}.}
\vspace{-2em}
\label{tab:task1-fixed-label}
\end{table}

\paragraph{Metrics}
Given a query $(C_r, e_q, r)$, with $C_r$ being the context (sentences containing events $e_q, e_t$ and their neighboring sentences) and $e_q$ as the source event, Task 1 is to generate a target event $e_t$ such that $r(e_q, e_t)$.
We format each query as ``In the context of $C$, what happens $r$ $e_q$?''.
We found formatting the query in natural language to be empirically better.
Let $\hat{e_t}$ be the system generated event.
We compare $e_t$ vs. $\hat{e_t}$ using \textsc{bleu}~\cite{papineni2002bleu}, \textsc{meteor}~\cite{denkowski2011meteor}, and \textsc{rouge}~\cite{lin2004rouge}\footnote{\citet{sharma2017nlgeval}, \url{https://github.com/Maluuba/nlg-eval}}, and measure the accuracy (\textsc{acc}) as the fraction of examples where $e_t = \hat{e_t}$.


\vspace{0.5em}
\noindent{\bf Results on \tggen }
The results are listed in Table~\ref{tab:task1-automated-metrics}.
Unsurprisingly, \gptz achieves high scores across the metrics showing that it is highly effective in generating correct events.
To test the generative capabilities of the models, we perform an ablation by removing the sentence containing the target event $e_t$ from $C_r$~(indicated with \textit{-C}). 
Removal of context causes a drop in performance for both \ours and \sts, showing that it is crucial for generating temporal events.
However, \gptz obtains higher relative gains with context present, indicating that it uses its large architecture and pre-training to use the context more efficiently.
\gptz also fares better as compared with \sts in terms of drop in performance for the out-of-domain \tbden dataset on metrics like accuracy ($-21\%$ vs. $-33\%$) and \textsc{bleu} ($-16\%$ vs. $-21\%$), indicating that pre-training makes helps \gptz in generalizing across the domains.

\noindent{\bf Human Evaluation}
To understand the nature of errors, we analyzed 100 randomly sampled incorrect generations.
For 53\% of the errors, \gptz generated a non-salient event which nevertheless had the correct temporal relation with the query. 
Interestingly, for 10\% of the events, we found that \gptz \textit{fixed} the label assigned by \caevo~(Table~\ref{tab:task1-fixed-label}), i.e., $e_t$ was incorrect but $\hat{e_t}$ was correct.

\subsection{Task 2: Graph Generation}

\begin{table}[ht]
\setlength{\tabcolsep}{0.25em}
\begin{tabular}{@{}llrrrrrrrrrrrr@{}}
\toprule
& Dataset & \bleu  & \textsc{mtr} & \textsc{rg} & \textsc{dot\%}  \\ \toprule
\sts   & \tggen  & 4.79 & 15.03 & 45.95  & 86.93 \\
\ours & \tggen  & \textbf{37.77} & \textbf{37.22} & \textbf{64.24}  & \textbf{94.47} \\ \midrule
\sts & \tbden  & 2.61  & 12.76  & 28.36  & 89.31  \\
\ours & \tbden  & \textbf{26.61} & \textbf{29.49}  & \textbf{49.26} & \textbf{92.37} \\ \bottomrule
\end{tabular}
\caption{Graph string metrics.}
\label{tab:task2-token}
\end{table}

\begin{table}[ht]
\setlength{\tabcolsep}{0.20em}
\small{
\begin{tabular}{@{}llrrrrrr@{}}
\toprule
& Dataset &  $v_P$ & $v_R$ & $v_{F_1}$ & $e_P$ & $e_R$ & $e_{F_1}$  \\ \toprule
\sts   & \tggen & 36.84 & 24.89 & 28.11 & 9.65 & 4.29 & 4.70 \\
\ours & \tggen  & \textbf{69.31}& \textbf{66.12}  & \textbf{66.34} & \textbf{27.95} & \textbf{25.89} & \textbf{25.22} \\ \midrule
\sts & \tbden  & 24.86 & 15.25 & 17.99 & 4.7 & 0.14 & 0.24 \\
\caevo & \tbden  & 37.53 & \textbf{79.83} & \textbf{48.96} & 7.95 & \textbf{14.62} & \textbf{8.96} \\
\ours & \tbden  & \textbf{45.96} & 48.44 & 44.97 &  \textbf{8.74} & 8.89 & 7.96 \\ \bottomrule
\end{tabular}
}
\caption{Graph semantic metrics.}
\label{tab:task2-semantics}
\end{table}

\noindent{\bf Metrics } Let $\V{G_i}(\V{V_i},\V{E_i})$ and $\V{\hat{G}_i}(\V{\hat{V}_i}, \V{\hat{E}_i})$ be the true and the generated graphs for an example $i$ in the test corpus.
Please recall that our proposed method generates a graph from a given document as a string in \dotlang.
Let $\S{y_i}$ and $\S{\hat{y_i}}$ be the string representations of the true and generated graphs.
We evaluate our generated graphs using three types of metrics:

\noindent 1. \textbf{Graph string metrics}: To compare $\S{y_i}$ vs. $\S{\hat{y_i}}$, we use \textsc{bleu}, \textsc{meteor}, and \textsc{rouge}, and also measure parse accuracy (\textsc{dot\%}) as the \% of generated graphs $\S{\hat{y_i}}$ which are valid \dotlang files.

\vspace{0.5em}
\noindent 2. \textbf{Graph structure metrics} To compare the structures of the graphs $\V{G_i}$ vs. $\V{\hat{G}_i}$, we use i) Graph edit distance (\ged)~\cite{abu2015exact} - the minimum numbers of edits required to transform the predicted graph to the true graph by addition/removal of an edge/node;
ii) Graph isomorphism (\textsc{iso})~\cite{cordella2001improved} -  a binary measure set to $1$ if the graphs are isomorphic (without considering the node or edge attributes);
iii) The average graph size $(|\V{V_i}| ,|\V{E_i}|, |\V{\hat{V}_i}|, |\V{\hat{E}_i}|)$ and the average degree~($\text{d}(\V{V})$).

\vspace{0.5em}
\noindent 3. \textbf{Graph semantic metrics: } We evaluate the node sets ($\V{V_i}$ vs. $\V{\hat{V}_i}$) and the edge sets ($\V{E_i}$ vs. $\V{\hat{E}_i}$) to compare the semantics of the true and generated graphs.
For every example $i$, we calculate node-set precision, recall, and $F_1$ score, and average them over the test set to obtain node precision ($v_P$), recall ($v_R$), and $F_1$ ($v_F$).
We evaluate the predicted edge set using temporal awareness~\cite{uzzaman2012interpreting,uzzaman2013semeval}.
For an example $i$, we calculate
$
    e_P^i = \frac{|\V{\hat{E}_i}^{-} \cap \V{E_i}^{+}|}{|\V{\hat{E}_i}^{-}|},
    e_R^i = \frac{|\V{\hat{E}_i}^{+} \cap \V{E_i}^{-}|}{|\V{E_i}^{-}|}
$
where symbol $+$ denotes the temporal transitive closure~\cite{allen1983maintaining} of the edge set.
Similarly, $-$ indicates the reduced edge set, obtained by removing all the edges that can be inferred from other edges transitively.
The $F_1$ score $e_{F_1}^i$ is the harmonic mean of $e_P^i$ and $e_R^i$, and these metrics are averaged over the test set to obtain the temporal awareness precision ($e_P$),  recall ($e_R$), and $F_1$ score ($e_{F_1}$).
Intuitively, the node metrics judge the quality of generated events in the graph, and the edge metrics evaluate the corresponding temporal relations.
\paragraph{Results}
Tables~\ref{tab:task2-token}, \ref{tab:task2-structure}, and \ref{tab:task2-semantics} present results for graph generation, and we discuss them next.

\begin{table}[ht]
\setlength{\tabcolsep}{0.30em}
\centering
\small{
\begin{tabular}{@{}llrrrrr@{}}
\toprule
       & Dataset & $|\V{V}|$ & $|\V{E}|$ & $\textit{d}(\V{V})$ & \ged  $\downarrow$  & \textsc{iso}  $\uparrow$  \\ \midrule
True   & \tggen  & 4.15  & 5.47 & 1.54 & 0 & 100 \\
\sts   & \tggen  & 2.24  & 2.23 &  1.12 & 6.09 & 32.49 \\
\ours  & \tggen  & \textbf{3.81}  & \textbf{4.60} & \textbf{1.40} &\textbf{ 2.62} & \textbf{41.66} \\ \midrule
True   & \tbden  & 4.39  & 6.12 & 2.02 & 0 & 100 \\
\sts   & \tbden  & 2.21  & 2.20 & 1.11 & 6.22 & 23.08 \\
\caevo & \tbden  & 10.73 & 17.68 & 2.76 & 18.68 & 11.11 \\
\ours  & \tbden  & \textbf{3.72}  & \textbf{4.65} & \textbf{1.75} & \textbf{4.05} & \textbf{24.00}  \\ \bottomrule
\end{tabular}
}
\caption{Graph structure metrics.}
\label{tab:task2-structure}
\end{table}

\paragraph{\gptz vs. \sts }
\gptz outperforms \sts on all the metrics by a large margin in both fine-tuned (\tggen) and zero-shot settings (\tbden).
\gptz generated graphs are closer to the true graphs in size and topology, as shown by lower edit distance and a higher rate of isomorphism in Table~\ref{tab:task2-structure}.
Both the systems achieve high parsing rates (\dotlang\%), with \ours generating valid \dotlang files 94.6\% of the time.
The high parsing rates are expected, as even simpler architectures like vanilla RNNs have been shown to generate syntactically valid complex structures like \LaTeX documents with ease~\cite{karpathy2015unreasonable}.

\paragraph{\gptz vs. \caevo}
We compare the graphs generated by \gptz with those extracted by~\caevo~\cite{chambers2014dense}\footnote{\url{https://github.com/nchambers/caevo}} from the \tbden documents.
We remove all the vague edges and the light verbs from the output of \caevo for a fair comparison.
Please recall that \caevo is the tool we used for creating the training data for our method.
Further, \caevo was trained using \tbden, while \gptz was not. 
Thus, \caevo forms an upper bound over the performance of \ours.
The results in Tables~\ref{tab:task2-semantics} and \ref{tab:task2-structure} show that despite these challenges, \ours performs strongly across a wide range of metrics, including \ged, \textsc{iso}, and temporal awareness.
Comparing the node-set metrics, we see that \ours leads \caevo by over eight precision points ($v_P$), but loses on recall ($v_R$) as \caevo extracts nearly every verb in the document as a potential event.
On temporal awareness (edge-metrics), \ours outperforms both \caevo and \sts in terms of average precision score $e_P$ and achieves a competitive $e_{F_1}$ score.
These results have an important implication: they show that our method can best or match a pipeline of specialized systems given reasonable amounts of training data for temporal graph extraction.
\caevo involves several sub-modules to perform part-of-speech tagging, dependency parsing, event extraction, and several statistical and rule-based systems for temporal extraction. 
In contrast, our method involves no hand-curated features, is trained end-to-end (single \ours), and can be easily scaled to new datasets.

\begin{table}
\centering
\small{\begin{tabular}{l} 
\toprule
\begin{tabular}[c]{@{}l@{}}\textbf{Top 10 Verbs:} found, killed, began, called, want,\\took, came, used, trying, asked \end{tabular}                            \\
\begin{tabular}[c]{@{}l@{}}\textbf{Randomly Sampled Verbs:} shooting, caused, \\accused, took, conceived, visit, vowing, play, \\withdraw, seems \end{tabular}  \\
\bottomrule
\end{tabular}}
\caption{Verbs in \gptz generated graphs.}
\label{tab:verbs-in-generated-graphs}
\end{table}

\paragraph{Node extraction and Edge Extraction}
The node-set metrics in Table~\ref{tab:task2-semantics} shows that \gptz avoids generating noisy events~(high $P$), and extracts salient events~(high $R$).
This is confirmed by manual analysis, done by randomly sampling 100 graphs from the \gptz generated graphs and isolating the main verb in each node~(Table~\ref{tab:verbs-in-generated-graphs}).
We provide several examples of generated graphs in the Appendix.
We note from Table~\ref{tab:task2-semantics} that the relative difference between the $e_{F_1}$ scores for \gptz and \sts (25.22 vs. 4.70) is larger than the relative difference between their $v_{F_1}$ scores~(66.34 vs. 28.11), showing that edge-extraction is the more challenging task which allows \gptz to take full advantage of its powerful architecture.
We also observe that edge extraction ($e_{F_1}$) is highly sensitive to node extraction ($v_{F_1}$); for \gptz, a 27\% drop in $v_{F_1}$ (66.34 on \tggen vs. 44.97 on \tbden) causes a 68\% drop in $e_{F_1}$ (25.22 on \tggen vs. 7.96 on \tbden).
As each node is connected to multiple edges on average (Table~\ref{tab:task2-structure}), missing a node during the generation process might lead to multiple edges being omitted, thus affecting edge extraction metrics disproportionately.

\begin{table}
\centering
\resizebox{\linewidth}{!}{%
\begin{tabular}{>{\hspace{0pt}}m{0.47\linewidth}>{\hspace{0pt}}m{0.30\linewidth}>{\hspace{0pt}}m{0.30\linewidth}} 
\toprule
Query ($C, e_q, r$) & $e_t$ & Explanation \\ \toprule
The suspected car bombings...turning busy streets...Which event happened before the suspected car bombings? & many cars drove & \textit{Plausible:} The passage mentions busy streets and car bombing. \\ \midrule
He...charged...killed one person. Which event happened after he was charged? & He was acquitted & \textit{Somewhat plausible:} An acquittal is a possible outcome of a trial. \\
\bottomrule
\end{tabular}
}
\caption{Sample open-ended questions and the answers $e_t$ generated by our system. Note that the answers generated by our system $e_t$ are complete event phrases (not just verbs).}
\label{tab:torque}
\end{table}

\subsection{Answering for Open-ended Questions}
A benefit of our approach of using a pre-trained language model is that it can be used to \textit{generate} an answer for open-ended temporal questions.
Recently, \citet{ning2020torque} introduced Torque, a temporal reading-comprehension dataset.
Several questions in Torque have no answers, as they concern a time scope not covered by the passage (the question is about events not mentioned in the passage).
We test the ability of our system for generating plausible answers for such questions out of the box (i.e., without training on Torque).
Given a (passage, question) pair, we create a query $(C, e_q, r)$, where $C$ is the passage, and $e_q$ and $r$ are the query event and temporal relation in the question.
We then use our \gptz based model for node-generation trained without context and generate an answer $e_t$ for the given query.
A human-judge rated the answers generated for 100 such questions for plausibility, rating each answer as being \textit{plausible}, \textit{somewhat plausible}, or \textit{incorrect}.
For each answer rated as either \textit{plausible} or \textit{somewhat plausible}, the human-judge wrote a short explanation to provide a rationale for the plausibility of the generated event. 
Out of the 100 questions, the human-judge rated 22 of the generated answers as plausible and ten as somewhat plausible, showing the promise of our method on this challenging task~(Table~\ref{tab:torque}).



%% file: conclusion-future.tex
\section{Conclusion and Future Work}
Current methods for generating event-level temporal graphs are developed with relatively small amounts of hand-labeled data. 
On the other hand, the possibility of using pre-trained language models for this task has not received sufficient attention.
This paper addresses this open challenge by first developing a data generation pipeline that uses existing \textsc{ie}/\textsc{nlp}/clustering techniques for automated acquisition of a large corpus of document-graph pairs, and by proposing a new formulation of the graph generation task as a sequence-to-sequence mapping task, allowing us to leverage and fine-tune pre-trained language models.
Our experiments strongly support the effectiveness of the proposed approach, which significantly outperforms strong baselines.
We plan to explore techniques for adapting large-scale language models on unseen domains and at multiple granularity levels in the future.
\section*{Acknowledgments}
Thanks to Nathanael Chambers and Dheeraj Rajagopal for the helpful discussion, and to the anonymous reviewers for their constructive feedback.
This material is based on research sponsored in part by the Air Force Research Laboratory under agreement number FA8750-19-2-0200. 
The U.S. Government is authorized to reproduce and distribute reprints for Governmental purposes notwithstanding any copyright notation thereon. 
The views and conclusions contained herein are those of the authors and should not be interpreted as necessarily representing the official policies or endorsements, either expressed or implied, of the Air Force Research Laboratory or the U.S. Government.

%% file: appendix.tex
\clearpage
\appendix

\section{Learning Event Communities Using Community Detection}
In this section, we provide the details on the community detection algorithm used by our method.
We define the temporal event communities to be a division of the temporal graph $\V{G}(\V{V}, \V{E})$ into sub-graphs $\V{G_1}(\V{V_1}, \V{E_1}), \V{G_2}(\V{V_2}, \V{E_2}), ..., \V{G_k}(\V{V_k}, \V{E_k})$ such that the events in a community (sub-graph) $\V{G_i}$ are more co-referential to each other as opposed to the other events in the temporal graph.
We use the undirected link between two events $e_j, e_i$ as a proxy for them being co-referential, and learn temporal event communities utilizing the concept of modularity, first introduced by~\cite{newman2004finding}.

Formally, let $\V{A}$ be the undirected adjacency matrix for a temporal graph $\V{G}(\V{V}, \V{E})$ such that $\V{A}(e_i, e_j) = 1$ if $e_i$ and $e_j$ are connected by a temporal relation, and 0 otherwise.
Further, let $\delta(e_i, e_j) = 1$ if events $e_i, e_j$ belong to the same temporal community, and $0$ otherwise.
For a given $\delta$, we denote the fraction of the edges that exist between events that belong to the same communities by $f_{same} = \frac{\sum_{e_i, e_j \in \V{E}}\V{A}(e_i, e_j)\delta(e_i, e_j)}{2|\V{E}|}$. 
Where the $2|\V{E}|$ in the denominator is due to the fact that $\V{A}$ treats $\V{G}$ as an undirected graph.
Let the popularity $p$ of an event $e_i$ be the number of events that are linked to it i.e. $p(e_i) = \sum_{e_j \in \V{E}}\V{A}(e_i, e_j)$.
The probability of randomly picking an event $e_i$ when sampled by popularity is $\frac{p(e_i)}{\sum_{e_j \in \V{E}}p(e_i)} = \frac{p(e_i)}{2|\V{E}|}$.
Thus, if edges are created randomly by sampling nodes by popularity $p$ of the nodes, the fraction of edges within the communities, $f_{random}$, is given by
\begin{align*}
    f_{random} &= \frac{\sum_{e_i, e_j \in \V{E}}p(e_i)p(e_j)\delta(e_i, e_j)}{2|\V{E}| * 2|\V{E}|}
\end{align*}

Finally, defining modularity, $Q$, to be $f_{same} - f_{random}$:
\begin{align*}
    Q &= \frac{1}{2|\V{E}|} * \sum_{e_i, e_j \in \V{E}}\V{A}(e_i, e_j) - \frac{p(e_i)p(e_j)\delta(e_i, e_j)}{2|\V{E}|}
\end{align*}
We want to learn community assignments $\delta$ that maximize $Q$.
A high $Q$ would promote $f_{same} > f_{random}$ and thereby encourage highly inter-connected event communities.
Calculating such $\delta$ directly is not tractable, since the complexity of such an operation would be at least exponential in the number of events~\cite{newman2004fast}.
We use the fast implementation provided by~\cite{clauset2004finding} for calculating event communities iteratively. The algorithm converges at $Q ~ 0.3$.
We use a similar approximation at test time: given a document $D$, we first break it down into sub-documents using \caevo and then feed each sub-document to our method. 

\begin{figure*}[ht]
\centering
{\includegraphics[height=0.25\textheight]{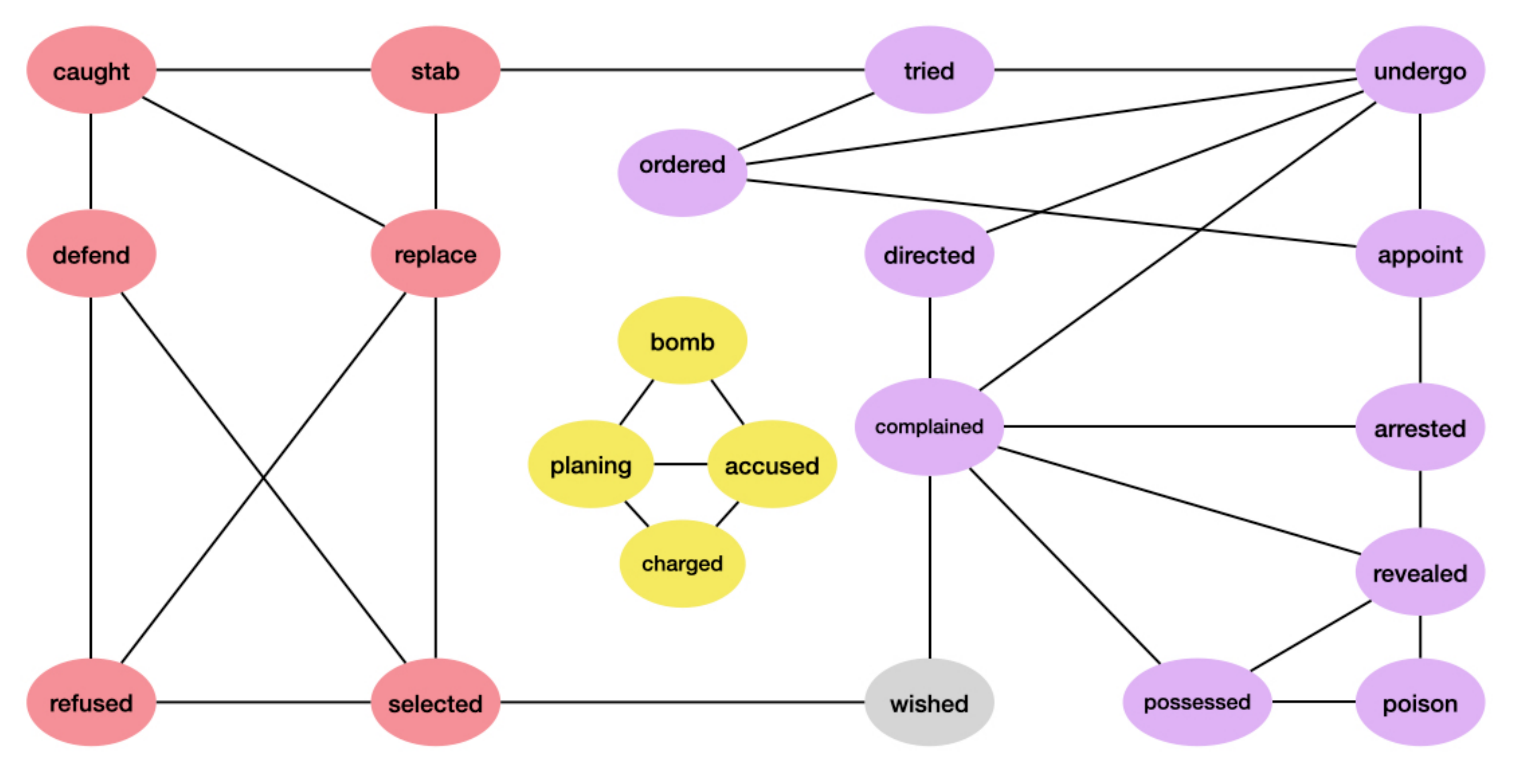}}
\caption{Event temporal graph and the extracted communities for a sample document. 
Each community is shown \textcolor{commred}{\textbf{in}} \textcolor{commyellow}{\textbf{different}} \textcolor{commblue}{\textbf{color}}.
The singleton nodes (\textcolor{darkgray}{\textbf{gray}}) are dropped. 
The nodes are only annotated with the verbs for brevity.
The edge labels and directions are not used for community detection.
}
\label{fig:event-communities}
\end{figure*}

\section{Using a smaller block size}
We found that the performance drops when using a block size of 300 and batch size of 2.
Table~\ref{tab:task2-token-bl300} presents the results.

\begin{table}[ht]
\centering
\begin{tabular}{rrrr}
\toprule
 \bleu  & \textsc{mtr} & \textsc{rg} & \textsc{dot\%}  \\ \toprule
 \textbf{25.01} & \textbf{27.95} & \textbf{60.99}  & \textbf{91.71}
\\ \bottomrule
\end{tabular}
\vspace{1.5em}

\begin{tabular}{@{}rrrrrr@{}}
\toprule
$v_P$ & $v_R$ & $v_{F_1}$ & $e_P$ & $e_R$ & $e_{F_1}$  \\ \toprule
 70.31 & 64.75 & 65.68 & 29.43 & 24.83 & 24.27  \\ \bottomrule
\end{tabular}

\caption{Results for \tggen using a block size of 300 and a block size of 2.}
\label{tab:task2-token-bl300}
\end{table}

\section{Masked Language Modeling Using Transformers}
In this section, we expand on the design of the transformer blocks.
For ease of reference, we re-iterate our training methodology.
We train a (separate) conditional language model to solve both the tasks.
Specifically, given a training corpus of the form $\{(\S{x_i}, \S{y_i})\}$, we aim to estimate the distribution $p_{\theta}(\S{y_i} \mid \S{x_i})$. 
Given a training example $(\S{x_i}, \S{y_i})$ we set $\S{u_i} = \S{x_i} \| \S{y_i}$\footnote{$\|$ denotes concatenation}.
$p_\theta(\S{u_i})$ can then be factorized as a sequence of auto-regressive conditional probabilities using the chain rule:  $p_\theta(\S{u_i}) = \prod_{k=1}^{n} p (u_{i,k} | \S{u_{i, <k}}) $, where $u_{i,k}$ denotes the $k^{th}$ token of the $i^{th}$ sequence, and $\S{u_{i, <k}}$ denotes the sequence of tokens $\{u_1, u_2, ..., u_{k - 1}\}$.
Language models are typically trained by minimizing a cross-entropy loss $-\text{log}p_\theta(\S{u_i})$ over each sequence $\S{u_i}$ in $\V{X}$.
However, the cross-entropy loss captures the joint distribution $p_\theta(\S{x_i}, \S{y_i})$, and is not aligned with our goal of learning conditional distribution $p_\theta(\S{y_i} | \S{x_i})$.
To circumvent this, we train our model by masking the loss terms corresponding to the input $\S{x_i}$, similar to \citet{Bosselut2019COMETCT}.
Let $\S{m_i}$ be a mask vector for each sequence $\S{u_i}$, set to $0$ for positions corresponding to $\S{x_i}$, and $1$ otherwise i.e. $m_{i, j} = 1$ $\text{if } j > |\S{x_i}|$, else 0.
We combine the mask vector with our factorization of $p_\theta(\S{u_i})$ to formulate a \textit{masked} language modeling loss, which is minimized over the training corpus $\V{X}$ to estimate the optimal $\theta$:
$$
     \mathcal{L}_{\textit{masked}}(\V{X}) = - \sum_{i=1}^{|\V{X}|} \sum_{j=1}^{|x_i| + |y_i|} m_{i, j}  * \text{log}(p_\theta (u_{i,j} | \S{u_{i, <j}}))
$$
Note that the formulation of masked loss is opaque to the underlying architecture, and can be implemented with a simple change to the loss function.
Intuitively, the model is optimized for only the output sequence $y_i$.
\subsection{Adapting \gptz for Masked Language Modeling}
\label{sec:gpt2mlm}
In practice, we use \gptz~\cite{radford2019language} based on transformer architecture~\cite{vaswani2017attention} for our implementation.
An input sequence $\S{u_i}$ of length $n$ is first embedded to a continuous representation denoted by $\S{u_i}^{(0)} \in \mathbb{R}^{nd}$.
$\S{u_i}^{(0)}$ is then passed through a series of $L$ \textit{transformer blocks} to obtain the output sequence $\S{u_i}^{(L)} \in \mathbb{R}^{nh}$.
Each transformer block~\cite{vaswani2017attention} consists of two operations: an auto-regressive version of the multiheaded self-attention~\cite{vaswani2017attention} operation (\textit{AutoRegMultiHead}) followed by a feed-forward operation (\textit{FFN}).
Each of these operations is surrounded by a residual connection~\cite{he2016deep} and followed by a layer normalization~\cite{ba2016layer} operation.
Denoting by $\S{u}^{(l - 1)}$ the input to the $l^{th}$ transformer block , the operations are in a transformer block are defined as follows:
\begin{align*}
\tilde{\S{u}}_{attn}^{l} &= \textit{AutoRegMultiHead}(\S{u}^{(l - 1)}) \\
\S{u}_{att}^{(l)} &= \textit{LayerNorm}(\tilde{\S{u}}_{att}^{(l)} + \S{u}^{(l - 1)}) \\
\tilde{\S{u}}_{ffn}^{(l)} &= \textit{FFN}(\S{u}_{att}^{(l)}) \\
\S{u}^{(l)} &= \textit{LayerNorm}(\tilde{\S{u}}_{ffn}^{(l)} + \S{u}_{att}^{(l)} ) \\
\end{align*}
Where $\textit{AutoRegMultiHead}$ is an auto-regressive version of the multiheaded self-attention~\cite{vaswani2017attention} that restricts the attention to the sequence seen so far (in accordance with the chain rule), and $\textit{FFN}$ is a feed-forward network (\textsc{mlp}).
After obtaining $\S{u_i}^{(L)}$, we set $p_{\phi}(\S{u_i}) = \text{softmax}(\S{u_i}^{(L)} * \V{W}_e)$, where $\mathbf{W}_e \in \mathbb{R}^{h|V|}$~($|V|$ is the size of the vocabulary).
Finally, we calculate the masked loss as  $\mathcal{L}(\S{u_i}) = \S{m_i}^{T} \odot \text{log} (p_{\phi}(\S{u_i}))$, and
the optimal $\phi$ is obtained by minimizing $\mathcal{L}_{\textit{masked}}(\V{X}) = - \sum_{i=1}^{|\V{X}|} \mathcal{L}(\S{u_i})$.

\section{Dataset Statistics}
Tables~\ref{tab:descriptors}, ~\ref{tab:verbcounts}, and~\ref{tab:relationtype} list various statistics calculated from the source data.
\begin{table}[ht]
\centering
\setlength{\tabcolsep}{0.25em}
\begin{tabular}{@{}lr@{}}
\toprule
Descriptor                            & \#Articles \\ \midrule
terrorism                             & 40909     \\
murders and attempted murders         & 25169     \\
united states international relations & 17761     \\
united states armament and defense    & 16785     \\
airlines and airplanes                & 16103     \\
world trade center (nyc)              & 15145     \\
demonstrations and riots              & 14477     \\
hijacking                             & 14472     \\
politics and government               & 6270      \\
bombs and explosives                  & 5607      \\ \bottomrule
\end{tabular}
\caption{Top Descriptors for the filtered Dataset. Note that each article is typically assigned more than one descriptor.}
\label{tab:descriptors}
\end{table}

\begin{table}[ht]
\setlength{\tabcolsep}{0.25em}
\centering
\begin{tabular}{@{}lrr@{}}
\toprule
Event verb & Raw frequency & \% Frequency \\ \midrule
said   & 647685 & 9.60  \\ 
say    & 57667  & 0.86 \\
had    & 47320  & 0.70  \\
killed & 43369  & 0.64 \\
told   & 42983  & 0.64 \\
found  & 41733  & 0.62 \\
made   & 40544  & 0.60  \\
war    & 35257  & 0.52 \\
get    & 30726  & 0.46 \\
make   & 29407  & 0.44 \\ \bottomrule
\end{tabular}
\caption{Most frequent events extracted by \caevo.}
\label{tab:verbcounts}
\end{table}

\begin{table}[ht]
\centering
\setlength{\tabcolsep}{0.25em}
\begin{tabular}{@{}lll@{}}
\toprule
Relation    & Raw Frequency & \% Frequency \\ \midrule
\before     & 2436201       & 54.51        \\
\after      & 1772071       & 39.65        \\
\isincluded & 131052        & 2.93         \\
\simul      & 112509        & 2.52         \\
\incld      & 17465         & 0.39         \\ \bottomrule
\end{tabular}
\caption{Relation Frequence in our Corpus}
\label{tab:relationtype}
\end{table}

\begin{table}[ht]
\centering
\begin{tabular}{@{}ll@{}}
\toprule
Relation    & Frequency \\ \midrule
\before     & 98715         \\
\after      & 68582         \\
\isincluded & 6179          \\
\simul      & 6209          \\
\incld      & 285           \\ \bottomrule
\end{tabular}
\caption{Edges in Generated Graphs: Top}
\label{tab:edges-in-generated-graphs}
\end{table}

\section{Examples}
\begin{figure*}[ht]
{\includegraphics[width=0.98\textwidth]{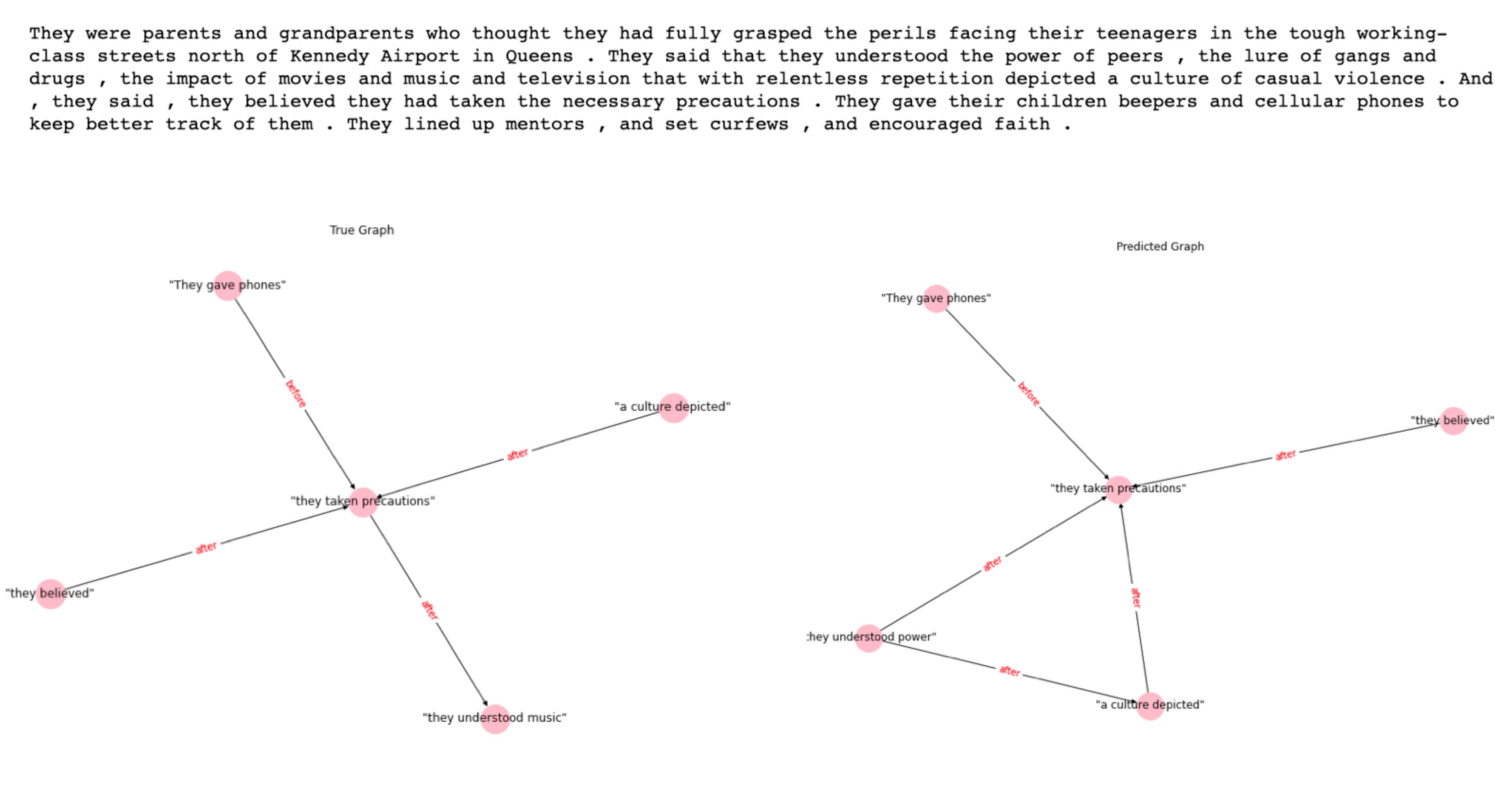}}
\caption{}
\label{fig:example1}
\end{figure*}

\begin{figure*}[ht]
{\includegraphics[width=0.98\textwidth]{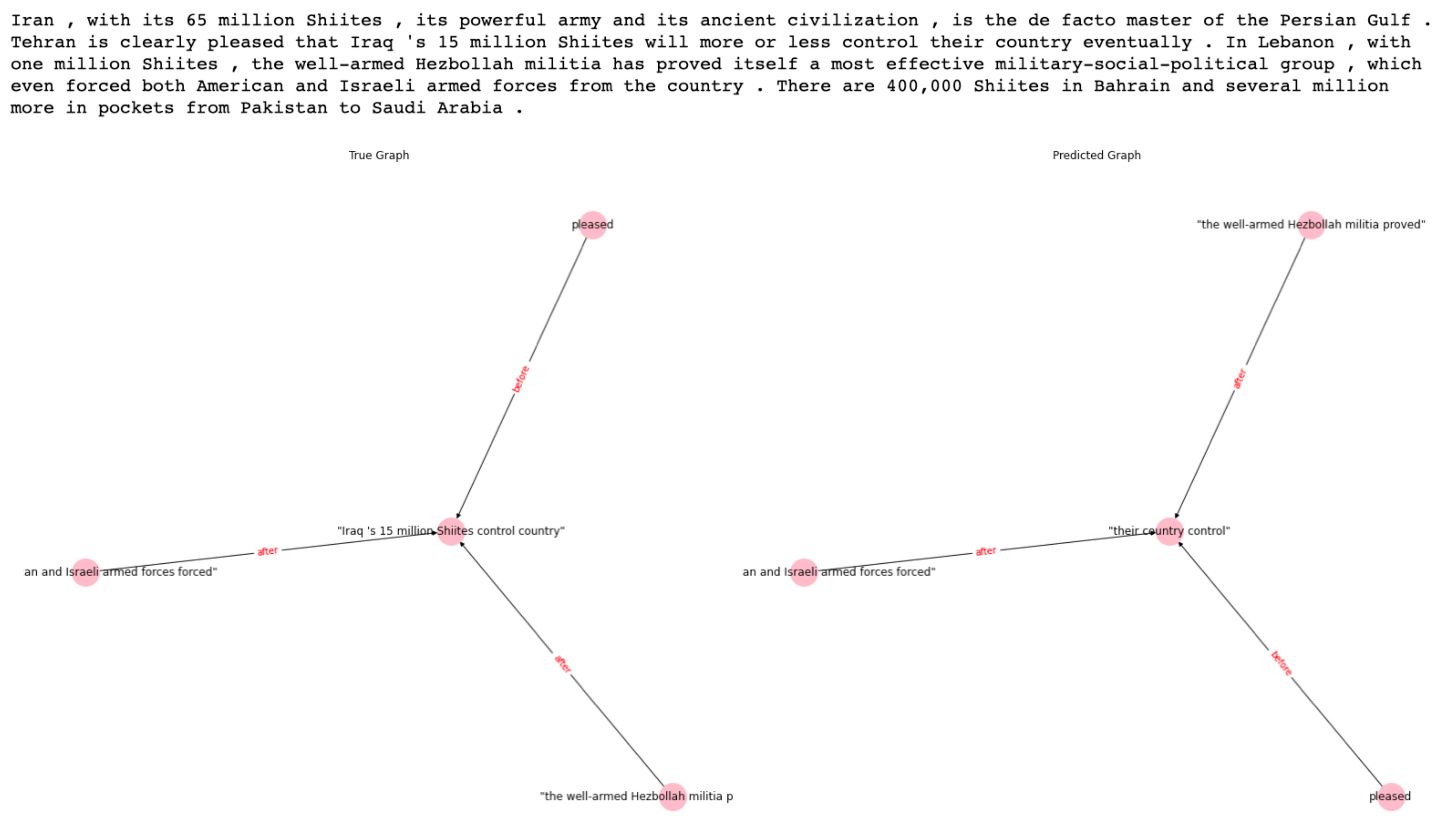}}
\caption{}
\label{fig:example2}
\end{figure*}

\begin{figure*}[ht]
{\includegraphics[width=0.98\textwidth]{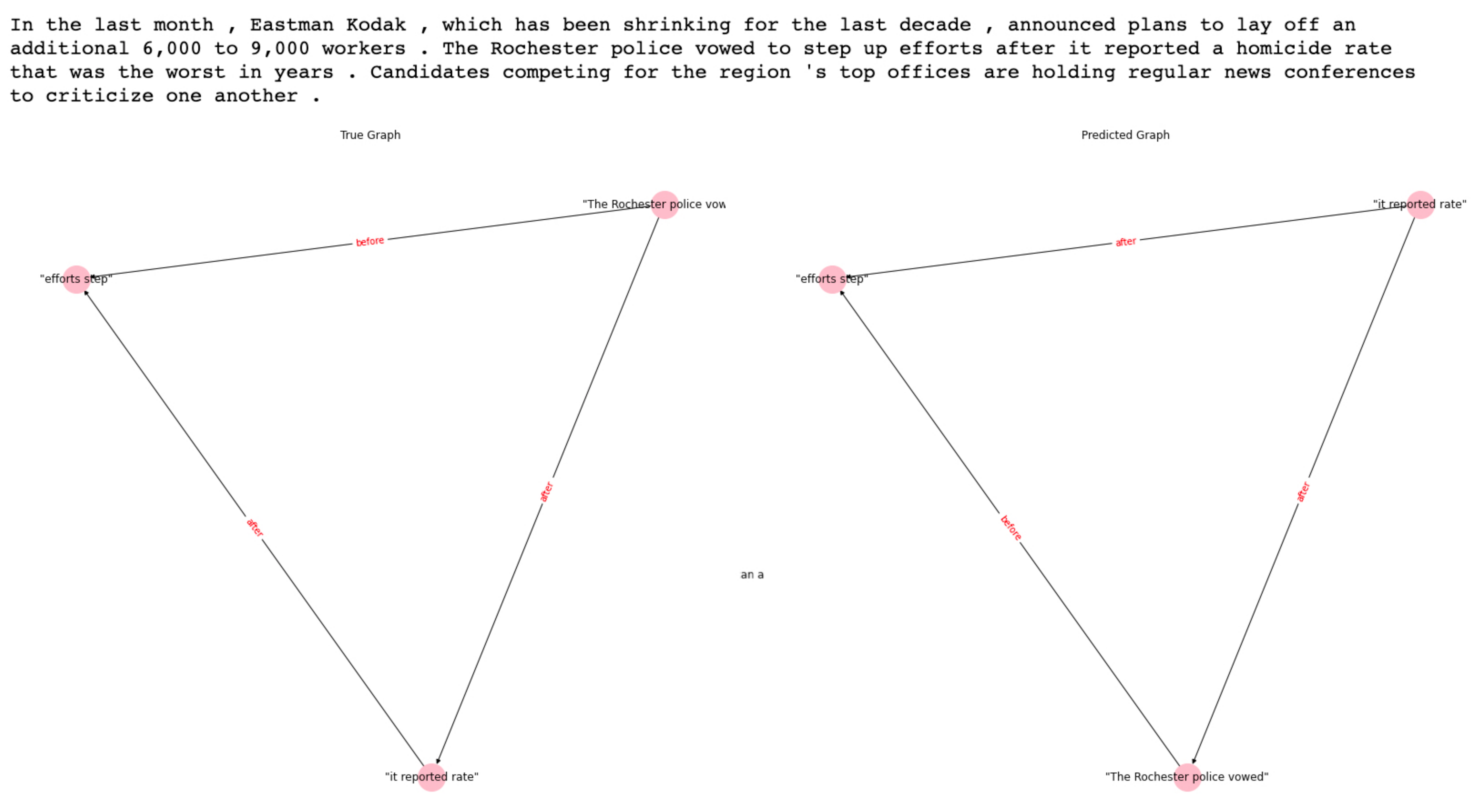}}
\caption{}
\label{fig:example3}
\end{figure*}

\begin{figure*}[ht]
{\includegraphics[width=0.98\textwidth]{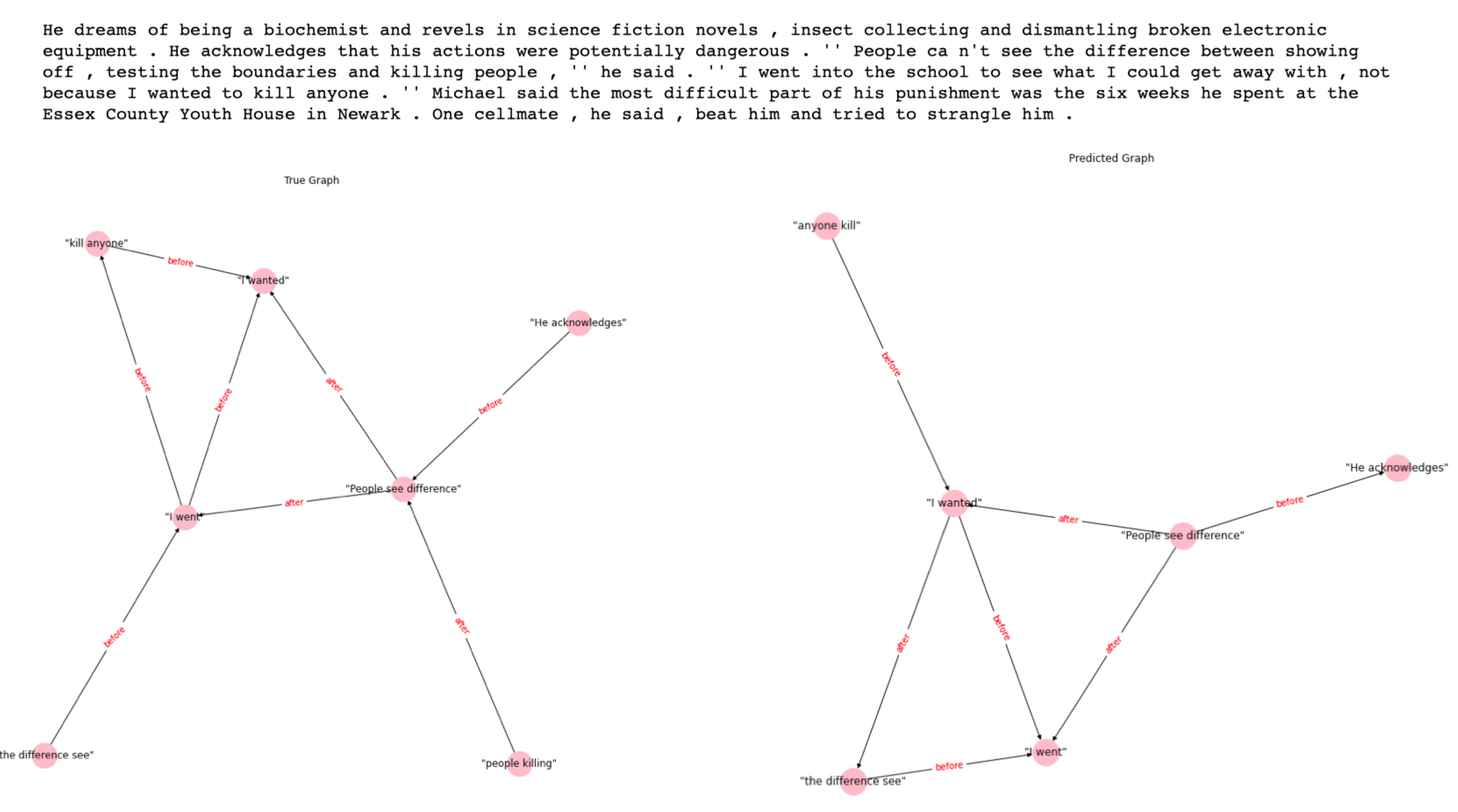}}
\caption{}
\label{fig:example4}
\end{figure*}

\begin{figure*}[ht]
{\includegraphics[width=0.98\textwidth]{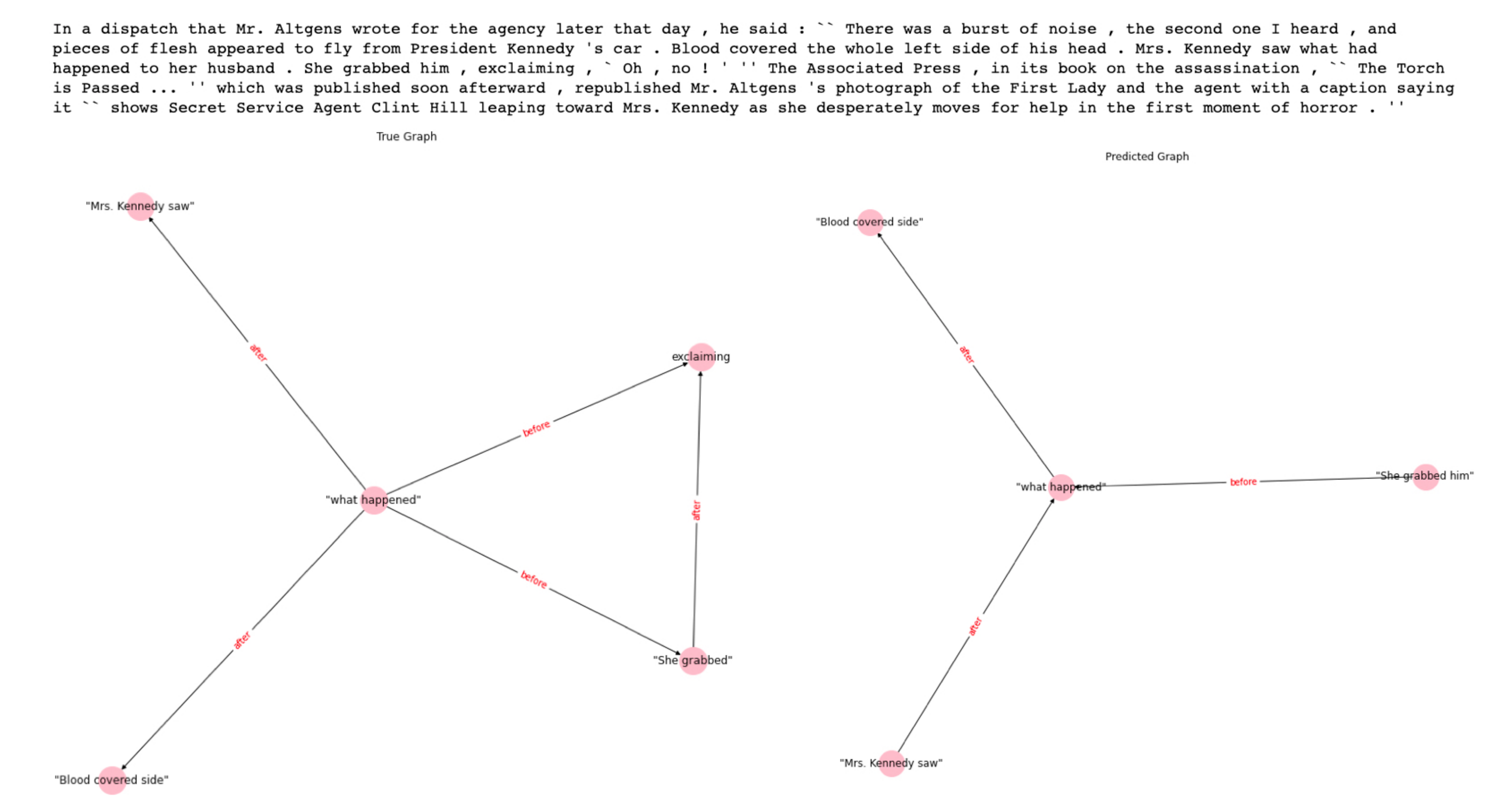}}
\caption{}
\label{fig:example5}
\end{figure*}

\begin{figure*}[ht]
{\includegraphics[width=0.98\textwidth]{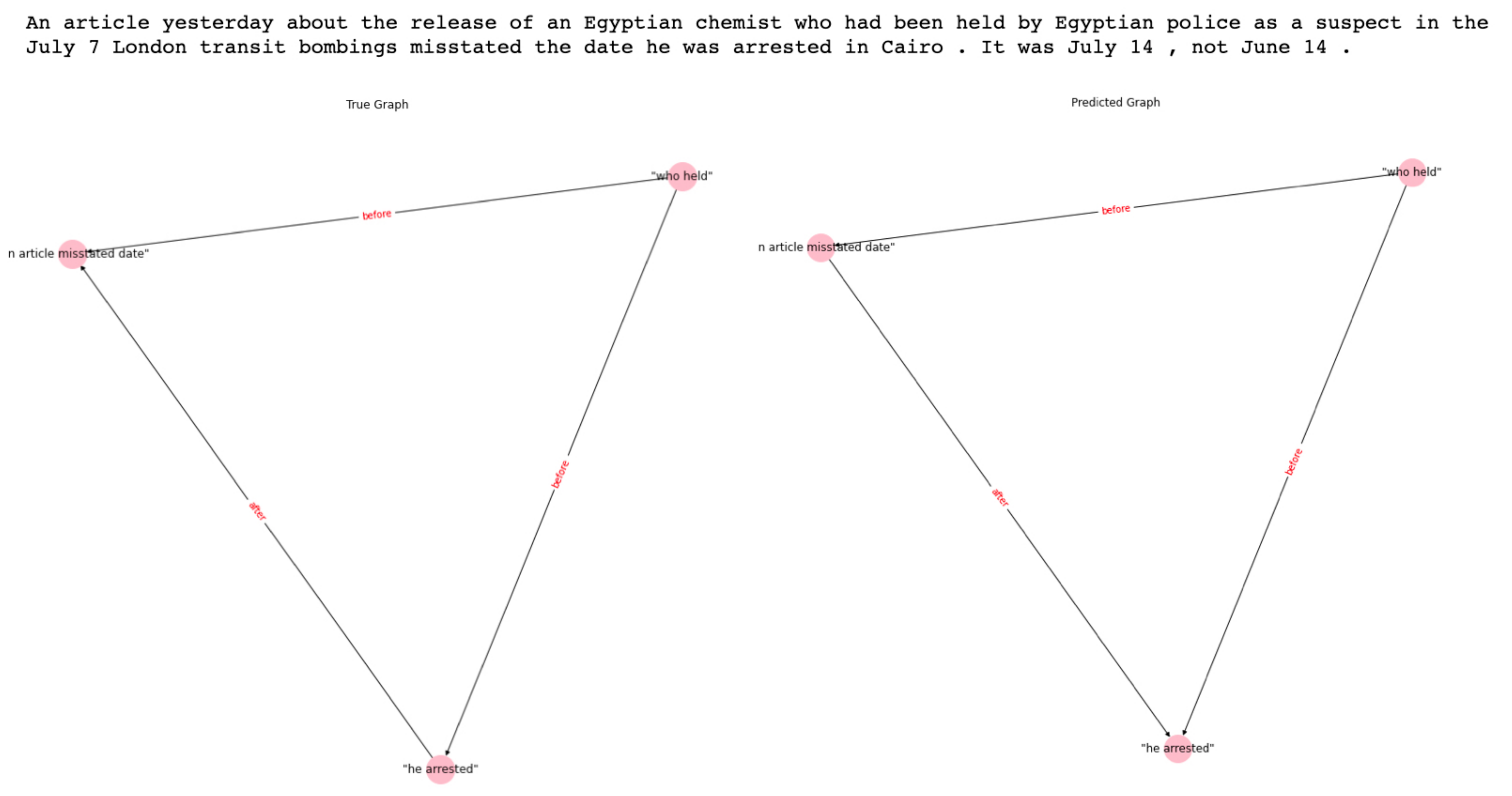}}
\caption{}
\label{fig:example6}
\end{figure*}

Figures~\ref{fig:example1}-\ref{fig:example6} show randomly picked examples from the test corpus. 
Each figure shows the text, the corresponding true graph, and the graph predicted by \gptz.

